# An Enhanced Harmony Search Method for Bangla Handwritten Character Recognition Using Region Sampling

Ritesh Sarkhel[1], Amit K Saha[1], Nibaran Das[1]

[1]*Computer Science and Engineering Department, Jadavpur University*
*Jadavpur, Kolkata 700032*
*Email: [sarkhelritesh, amitksweb, nibaran]@gmail.com*

***Abstract-*** **Identification of minimum number of local regions of a handwritten character image, containing well-defined discriminating features which are sufficient for a minimal but complete description of the character is a challenging task. A new region selection technique based on the idea of an enhanced Harmony Search methodology has been proposed here. The powerful framework of Harmony Search has been utilized to search the region space and detect only the most informative regions for correctly recognizing the handwritten character. The proposed method has been tested on handwritten samples of Bangla Basic, Compound and mixed (Basic and Compound characters) characters separately with SVM based classifier using a longest run based feature-set obtained from the image sub-regions formed by a CG based quad-tree partitioning approach. Applying this methodology on the above mentioned three types of datasets, respectively 43.75%, 12.5% and 37.5% gains have been achieved in terms of region reduction and 2.3%, 0.6% and 1.2% gains have been achieved in terms of recognition accuracy. The results show a sizeable reduction in the minimal number of descriptive regions as well a significant increase in recognition accuracy for all the datasets using the proposed technique. Thus the time and cost related to feature extraction is decreased without dampening the corresponding recognition accuracy.**

## I. INTRODUCTION

Optical Character Recognition (OCR) for handwritten characters is an active area of research for researchers all around the globe [1]. Motivation behind this is its large scope of applications; but the success of commercially available OCR could not be extended to handwritten characters as various writing styles make it quite difficult to identify the discriminating features of the characters itself. In spite of the huge popularity[2] of Bangla script, OCR of complete Bangla alphabet of handwritten characters has not received much attention from researchers until very recently. Due to numerous writing styles, huge and complex alphabet and presence of abundant Compound characters, Bangla script poses a challenge to the researchers.

One of the most common approaches of feature extraction in handwritten character recognition is to segment the sample image into several regions, extract the local features after pinpointing the minimum number of regions which are most informative in discriminating the character from others[3]. To obtain the local feature-set, different techniques are present in the literature [4][5], but none of them ensures optimal success rate [5]. Heuristics methods are applied to search the optimal, most informative regions out of all possible local regions. Several nature inspired meta-heuristics algorithms such as Genetic Algorithm[5], Artificial Bee Colony [6], Bacterial Foraging[7] etc. are used very recently. From this perspective, a new region sampling method has been introduced here for recognition of handwritten Bangla characters. The proposed methodology has been tested on the databases of Bangla Basic, Compound and a mixed dataset of both Basic and Compound characters.

| | | | | |
|---|---|---|---|---|
| *Sample Character Image* | 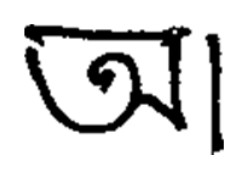 | 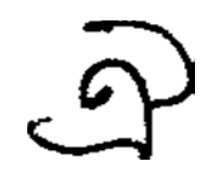 | 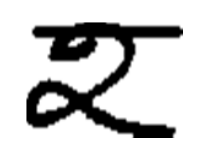 | 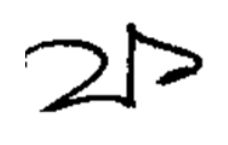 |
| *Character class* | আ | ঐ | হ | ফ |

**Fig 1**: Sample of Bangla Basic characters

There are several works mentioned in the literature to segment the image into several fixed sized windows or sub-regions and generates the local feature-set[3]. In [5] Das et al. used region sampling effectively and provided a comparative analysis between three meta-heuristics algorithm GA(genetic Algorithm), SA(Simulated Annealing) and HC(Hill Climbing) to search the region space for optimal number of most informative regions. Roy et al.[6] used Artificial Bee Colony Optimization method for sampling local regions. In the present work, an enhanced Harmony Search is used to sample local regions where shapes of the characters differ most.

## II. OVERVIEW OF BASIC HARMONY SEARCH ALGORITHM

Harmony Search algorithm (HS) is one of the most popular nature inspired, derivation free meta-heuristics algorithm. Proposed by Geem et al. [8]. It imitates a musician's journey towards finding a better state of harmony (as shown in Fig 2). HS has gained a lot of its popularity during recent years as it has been successfully used to solve many real life optimization problems such as Water network design[9], vehicle routing[10], pipe network design[11] etc. A brief overview of basic HS algorithm is presented in Fig. 3.

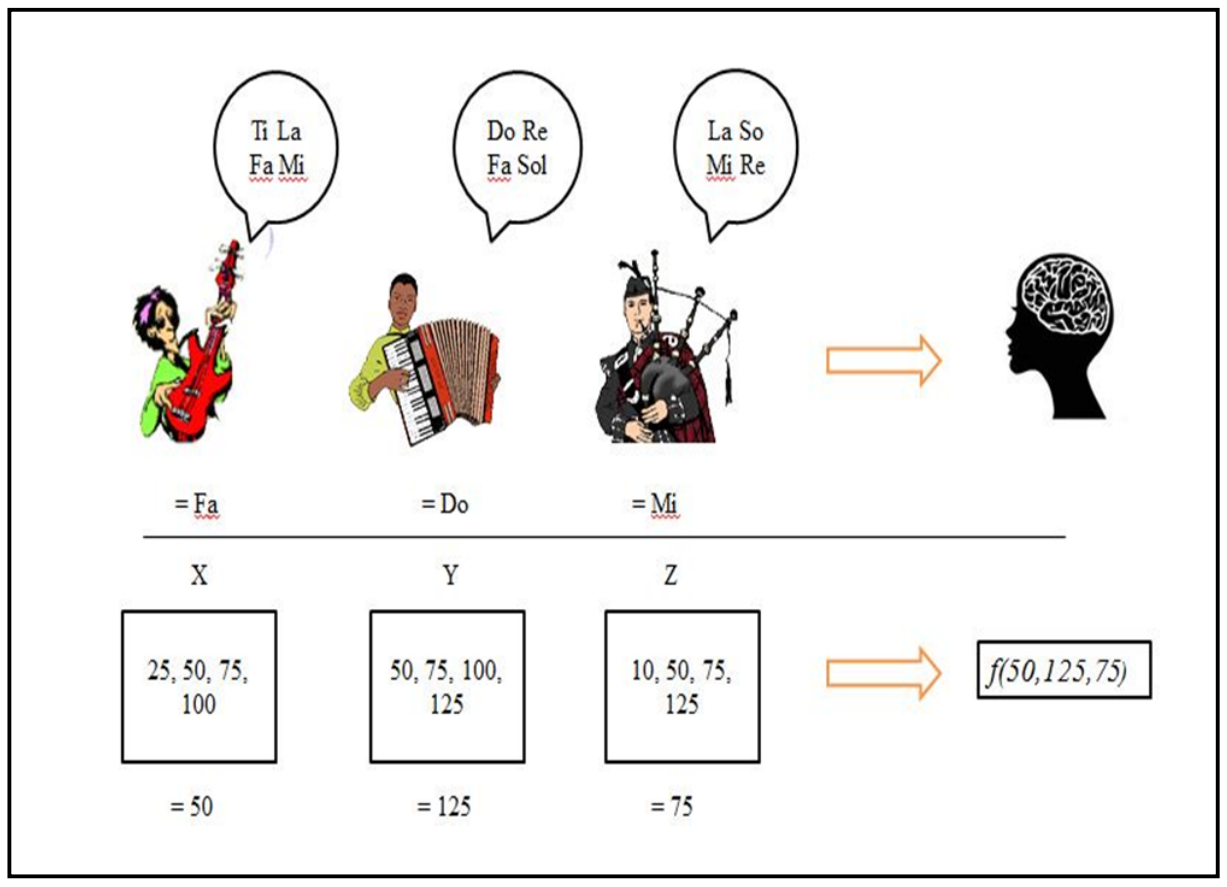

**Fig 2**: Analogy of musical improvisation and functional optimization in Harmony Search

## III. THE PRESENT WORK

The objectives of the present work are three-fold: (a) designing a region sampling strategy to select the regions of the image containing the most discriminating features describing the character, (b) evaluating the performance of the proposed method on a dataset containing Bangla Basic characters, Bangla Compound characters and a randomly mixed dataset containing both Bangla Basic and Compound characters, (c) performing a comparative analysis of the proposed method with the Basic HS algorithm for all the datasets. Fig. 4 and Fig. 5 show the block diagram of the region selection strategy of the proposed method.

### A. *Definitions and notations*

Let $I_{H,W}$ denotes a 2D array representing a digital image M of dimension $H \times W$ such that $I_{H,W} = \{f(i,j) | 0 \le i \le H-1 \text{ and } 0 \le j \le W-1\}$. f(i,j) denotes the intensity of the pixel at (i,j). Clearly, for a binary image f(i,j) ε {0,1}. A region $R_k$ is defined as a bounding rectangle $R_k(i_k^{TL}, i_k^{BR}, j_k^{TL}, j_k^{BR})$, such that $(i_k^{TL}, j_k^{TL})$ denotes the pixel at the top left corner and $(i_k^{BR}, j_k^{BR})$ denotes the pixel at the bottom right corner, where $0 \le i_k^{TL}, i_k^{BR} \le H-1$ and $0 \le j_k^{TL}, j_k^{BR} \le W-1$ for all $R_k \subseteq I_{H,W}$. A 2D digital image can be defined a set of its constituent regions, say S such that S = $\{R_1, R_2 \ldots\ldots, R_n\}$. The objective is to find a subset $S_i$ of S i.e. $S_i \subseteq S$ where $|S_i|$ is minimum among all such subsets of S which maximally increases the recognition accuracy of the image.

1. *Initialize the algorithm parameters: Harmony Memory Size (HMS), Harmony Memory Consideration Rate (HMCR), Pitch Adjustment Rate (PAR) and maximum number of iterations (NI).*
2. *Initialize the initial population of harmony memory (HM) with random harmonies, say $h_i$ such that $h_i$ ε $[lb_i, ub_i]$.*
3. *If $r_1 \le HMCR$; where $r_1$=rand () s.t. $r_1$ ε U(0, 1)*
   *3.1 Select a harmony randomly from HM, say H.*
   *3.2 If $r_2 \le PAR$; where $r_2$ = rand () s.t. $r_2$ ε U(0, 1)*
   *3.2.1 $H_{new} = H \pm r*BW$; where r = rand () s.t. r ε (0, 1)*
4. *Compare $H_{new}$ and the worst harmony of HM, say $H_{worst}$ in terms their corresponding objective function values f ($H_{new}$) and f($H_{worst}$) respectively; where f(.) is the fitness function.*
5. *If $H_{new}$ is fitter than $H_{worst}$, it is replaced by $H_{new}$ in the harmony memory.*
6. *Steps 3 to 5 is repeated until some pre-defined termination criterion is met or the number of iterations has reached to its maximum value NI.*

**Fig 3**: A brief overview of basic harmony search algorithm

### B. *Dataset of the experiment*

The proposed method has been tested on the datasets of Bangla Basic[12], Bangla Compound[13] and randomly mixed dataset of both Bangla Basic and Compound characters. It is worthy to mention here that Bangla alphabet contains 50 Basic characters; out of these 11 are vowels and 39 characters are consonants. Apart from these it is also enriched with more than 334 compound characters [5]. The datasets used in the experiment are developed at CMATER lab, Jadavpur University, Kolkata.

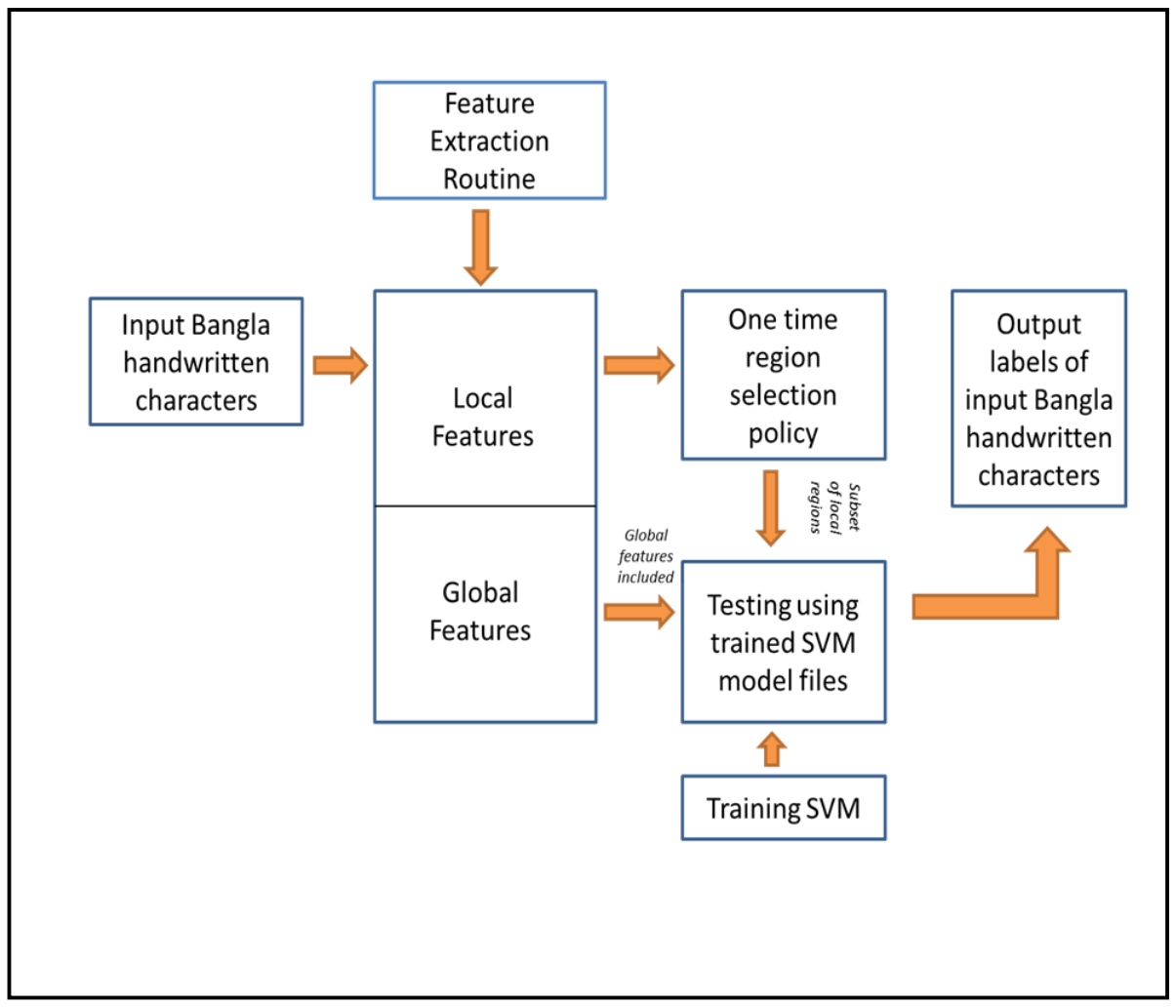

**Fig 4:** Block diagram of the proposed system

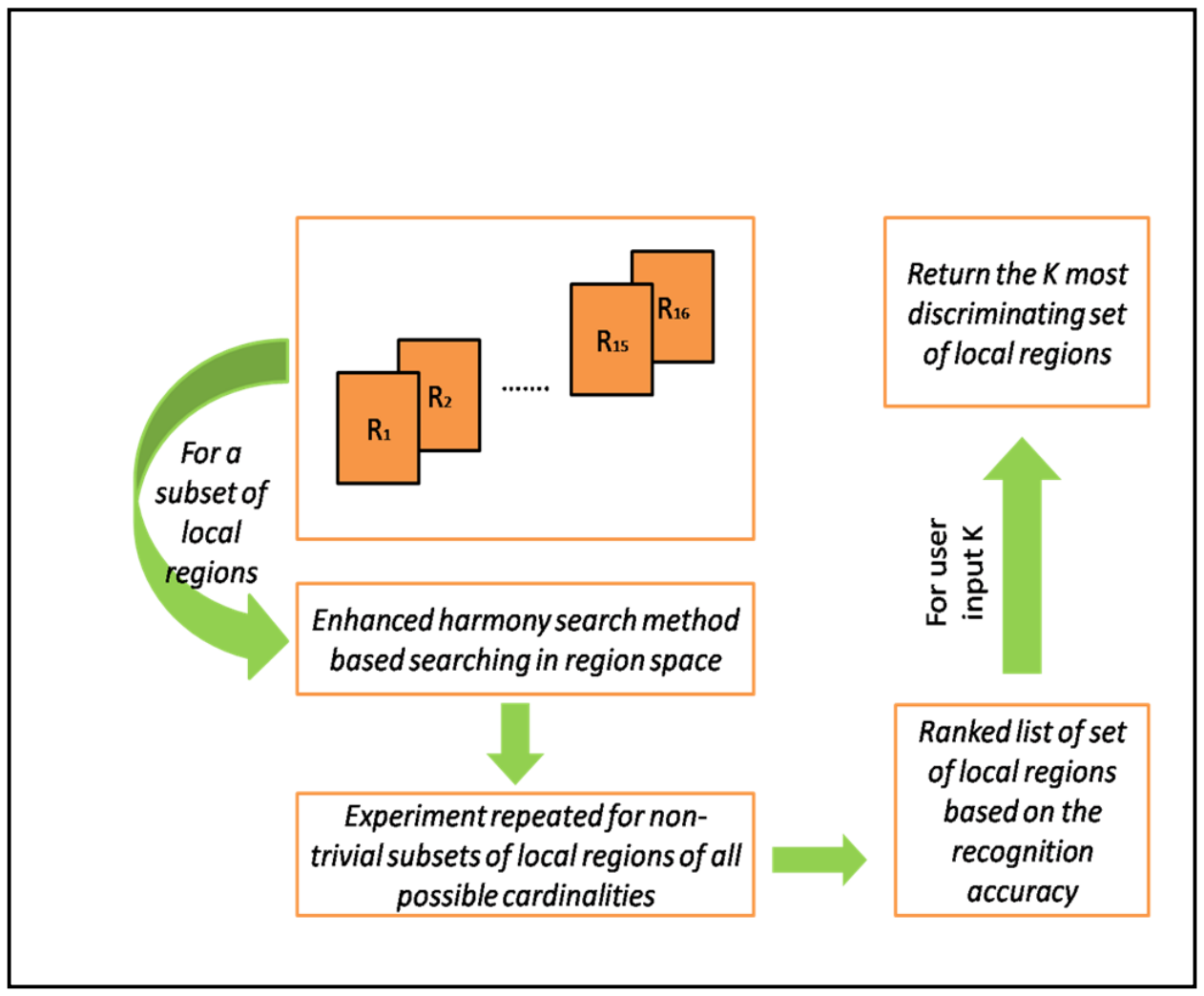

**Fig 5:** One time region selection policy

### C. Design of the feature set

The feature set used in the experiment consists of longest run features[5] computed along 4 axes, horizontal axis, vertical axis and two major diagonal axes respectively. The features are extracted from the sub-regions, at levels 0, 1 and 2 of the quad-tree, partitioned based on CG as suggested by Basu et al [14][18]. Hence total number of features in the feature set is ( $1 \times 4 + 4 \times 4 + 16 \times 4 = 84$). Features extracted from levels 0 and 1 of the quad-tree comprise the global feature-set and the features extracted from level 2 comprise the local feature-set of the experiment. Fig. 6 provides a brief overview of the feature extraction methods that generate the initial feature-set for this experimental setup.

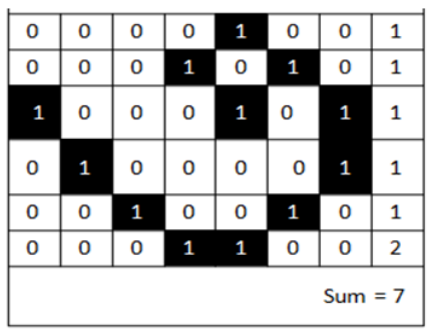

*(a) Row wise longest-run based feature calculation of a portion of binary image enclosed by rectangular bounds*

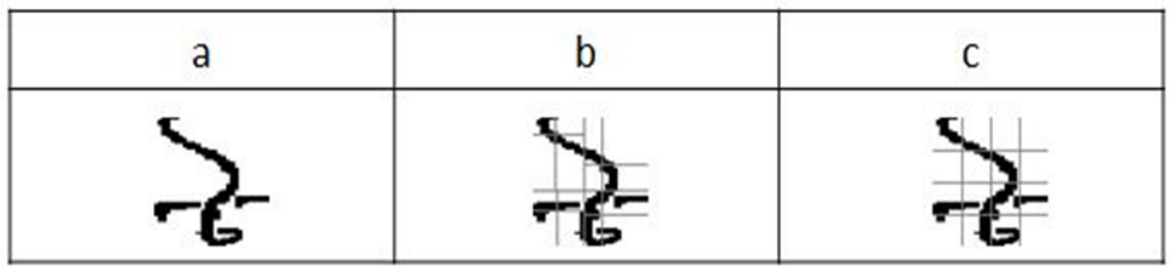

***(b)** Different partitioning techniques: (i) Sample handwritten character (ii) 2-level CG based quad tree partitioning (iii) Equal partitioning*

**Fig 6:** Feature extraction technique used in the present work

### D. Region sampling methodology

Let i denotes the quad-tree based image partition levels. As described earlier $i \varepsilon \{0, 1, 2\}$. $S_i$ denotes the set of regions extracted from $i^{th}$ level, $S_i = \{R_{ij} \mid j \varepsilon \{0, 1 \ldots, 4^i - 1\}\}$. $V_{ij}$ denotes the set of feature values extracted from region $R_{ij}$. M $= \cup_j^i R_{ij}$ denotes the sample image. Longest run based features are extracted from M, for each sub-region generated by the partition. Hence $V_{ij} = \{F^R_{ij}, F^C_{ij}, F^{D1}_{ij}, F^{D2}_{ij}\}$. Fitness value for the set of regions $C \subseteq M$ is the recognition accuracy of corresponding Bangla handwritten character by SVM classifier, using only the features extracted from the set of regions $C_G$, where $C_G = C \cup G$; G is the set of global regions. Therefore if f(.) denotes the fitness function, fitness value corresponding to the set of regions C will be $f(C_G)$.

#### (a) Initialization:

The algorithm is initialized with an empty set of regions **B**, i.e. **B** = Φ. Upon termination, **B** contains the best set of regions returned by the algorithm. The parameters of the algorithm *Harmony Memory Consideration Rate* (HMCR), *Pitch Adjustment Rate* (PAR) and *Bandwidth* (BW) are set as advised by Geem et al. for best performance in [15]. *Harmony Memory Size* (HMS) is total number of local regions of the image. Only local regions are sampled for further optimization, hence the initial harmony memory (HM) is comprised of only the set of regions extracted from the $2^{nd}$ level of quad-tree partitioning. A roulette wheel is created such that each sector corresponds to a candidate region of harmony memory and is equivalent to its fitness value.

#### (b) Selection of most informative set of regions

One of the objectives of the present work is to identify the minimal set of most discriminating regions by heuristically searching the region space.

#### (c) Termination Criterion

The algorithm is terminated when the experiment has successfully produced 25 generations for each possible size of non-trivial local region subset. Hence maximum number of iterations (NI) of the algorithm is 25.

The methodology of selecting minimum number of most informative set of regions that is used in the present work is described as follows.

**Algorithm 1:** An enhanced harmony search based region sampling methodology

***Input:*** *Initial feature-set extracted from the sample Bangla handwritten character*

***Output:*** *Maximum recognition accuracy achieved by the method and minimum number of sub-regions used to identify the Bangla handwritten character.*

---

*Initialize the algorithm parameters: HMS, HMCR, PAR, NI and maxSuccessrate.*

***Begin***

***for*** *k = HMS-1 to 1 do*

*/* HMS denotes the total number of local regions */*

*{*

*Spin the roulette wheel k times to populate the harmony memory with the initial population.*

***for*** *i = 1 to NI do*

***{***

*Generate a random number $r_1$ with uniform distribution between (0, 1).*

***if*** $r_1 \leq HMCR$

*{*

*Spin the roulette-wheel and select a region from the HM, say $R_i$.*

*/* $R_i$ is the candidate solution in this iteration */*

*Generate a random number $r_2$ with uniform distribution between (0, 1).*

***if*** $r_2 \leq PAR$

*{*

*Generate a ranked list L.*

*/* L contains all the regions comprising HM, sorted on their fitness values in a descending order. Ranks are given from 1 to |L|. */*

***if*** *($R_i$ is the $j^{th}$ entry in L)*

*{*

*Replace the last candidate of L with the $(j\text{-}1)^{th}$ entry of L.*

***}***

***}***

***}***

***else***

***{***

$S^c = M - HM$.

*Randomly select a region $R_{new}$ from the set of regions $S^c$ not included in HM.*

***if*** *($R_{new}$ is fitter than the last entry in the ranked list L)*

*/*Fitness value is the recognition accuracy of the character by SVM classifier */*

*{*

*Replace the last entry of L with $R_{new}$.*

*}*

*}*

*Find the best fitness value from the harmony memory, say $f(L_{best})$*

*/*Fitness value is the recognition accuracy of the character by SVM classifier */*

***if (*** *$f(L_{best})$ > maxSuccessrate)*

*/* maxSuccessrate is initialized with recognition accuracy of the initial feature-set by SVM classifier */*

*{*

*maxSuccessrate = $f(L_{best})$*

*}*

$B = \Phi$

$B = B \cup HM$

*}*

*}*

*/* B contains the minimum set of local regions used to identify the handwritten Bangla character and maxSuccessrate contains the maximum recognition accuracy achieved by the proposed method*/*

***End***

---

### *(d) Analysis of algorithm*

This method uses the framework of HS algorithm with some enhancements to further improve its performance. A roulette wheel selection method is used for both initial population generation and memory selection. This gives some form of control over the quality of harmony selected during the improvisation of the algorithm. Iterative decrement of harmony memory size guarantees that minimal numbers of most discriminating regions are being considered. Heuristics of the proposed method reduces the time complexity of the algorithm to $O(c + 4^{L-1})$ from $O(c + 2^{4^{L-1}})$ which would have been required for exhaustively searching the region space for most discriminative regions.

## IV. RESULTS OF THE EXPERIMENT

The integrated system design shown in Fig. 4.a is implemented for our experimental setup. For classification purposes, support vector machine or SVM[16] with RBF kernel is used. The gamma and nu values of RBF kernel are set empirically for the experiment. Among many implementations of SVM present in the literature, LIBSVM[17], an open source SVM tool is used here. As discussed earlier in the present work, experiments are performed over 3 different datasets: (a) Bangla Basic character set (b) Bangla Compound character set and (c) Randomly mixed set of both Bangla Basic and Compound characters. Finally, a comparative analysis is done to measure the performance of our proposed method, based on the metrics of recognition accuracy of the classifier and minimal number of discriminative regions used. Experimental results are shown in Table 1, Fig 7.a and Fig 7.b.

Table 1 shows the recognition accuracies achieved by all the methods. Figure 7.a and 7.b provide a comparison of the minimum number of most discriminative regions used for identifying the character. From the data collected, it can be observed that the proposed method achieves a significant 2.3% increment in recognition accuracy with 43.75% less number of discriminating regions for Bangla Basic characters, 0.6% increment in accuracy with 12.5% less number of discriminating regions for Bangla Compound characters and finally 1.2% increment in recognition accuracy with 37.5% decrease in number of discriminating regions is observed for dataset of mixed Bangla Basic and Compound characters.

**TABLE 1**: COMPARISON OF RECOGNITION ACCURACY ON TEST DATASETS

| Dataset | Present Work | Basic Harmony Search based region-sampling | Without region-sampling |
|---|---|---|---|
| Bangla Basic character-set | 86.5330% | 84.7212% | 84.5750% |
| Bangla Compound character-set | 78.3803% | 77.9108% | 77.8991% |
| Mixed character-set | 72.8753% | 72.0130% | 72.0093% |

**TABLE 2**: COMPARISON OF AVERAGE TIME TAKEN FOR EACH CHARACTER CLASSIFICATION OF TEST SET

| Dataset | Present Work | Basic Harmony Search based region-sampling | Without region-sampling |
|---|---|---|---|
| Bangla Basic character-set | 0.002341 | 0.002727 | 0.003008 |
| Bangla Compound character-set | 0.013203 | 0.013203 | 0.013734 |
| Mixed character-set | 0.019562 | 0.019708 | 0.021095 |

The reason behind this may be due to large number of classes and training samples used, uniquely different character shapes, resulting in different treatment to creation of the dataset. On other hand, relative decrease in number of informative regions is least for Bangla Compound characters. This might be due to the fact that shapes of Bangla Compound characters are very intricate and complex; sometimes the only thing to distinguish between two different Compound characters is a period or a small line, as discussed by Das et al.[18]. So it has the least number of region rejections and least increment in recognition accuracy.

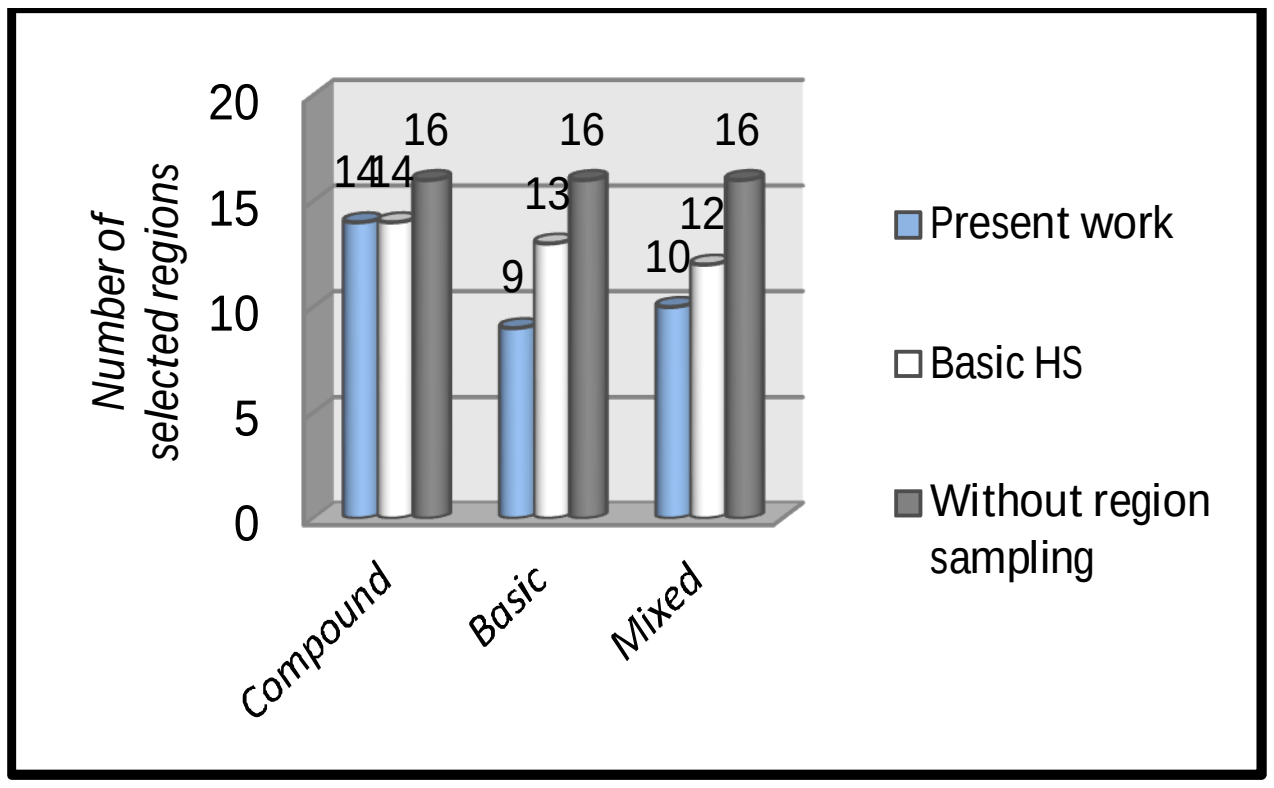

**Fig 7.a**: Comparison of selected regions

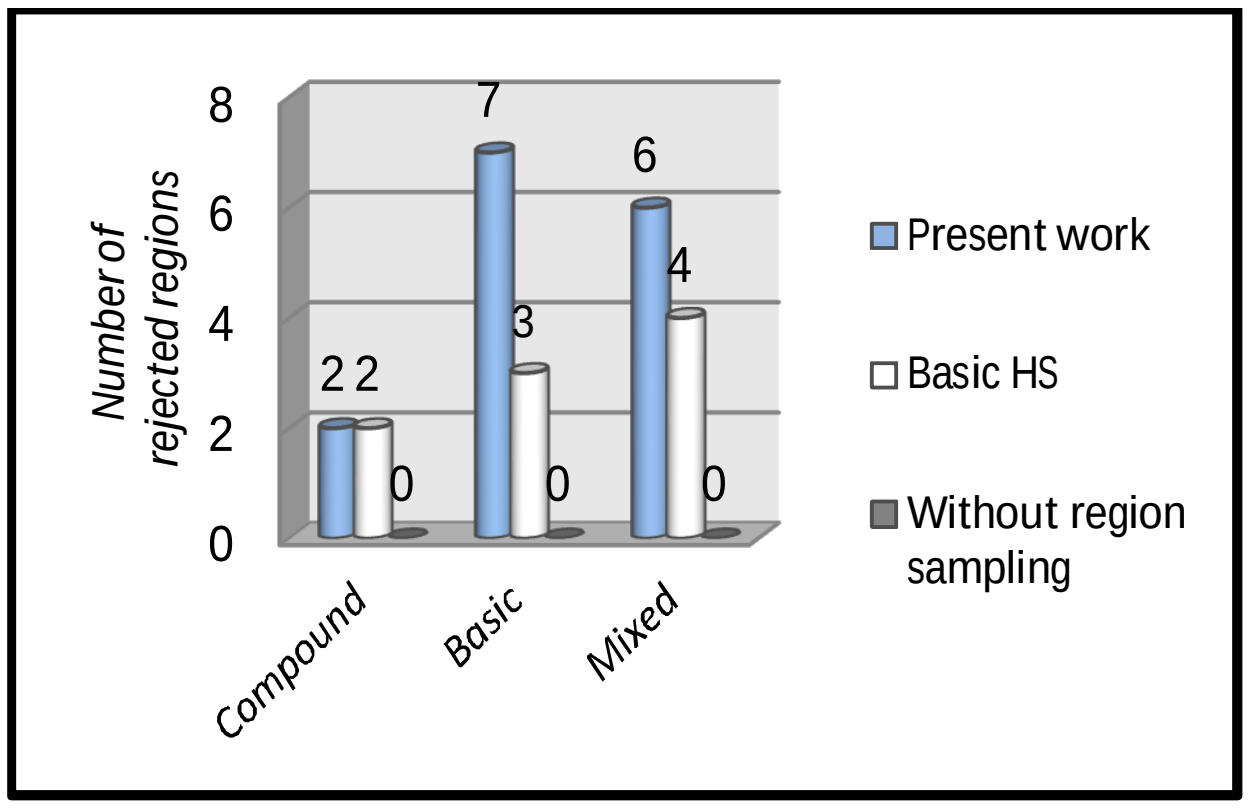

**Fig 7.b**: Comparison of rejected regions

Table 2 shows the comparison of average time (in seconds) needed to predict a test sample by SVM classifier. The tests have been performed on an Intel$^{R}$ Core$^{TM}$ i3-3110M processor, with clock frequency of 2.4 GHz and 2.00 GB RAM. From the results in Table 2 it is clear that the proposed method performs faster classification of test samples than its contemporaries. This may be attributed to better representation of a character by using lesser number of regions and using only the most informative regions. Fig. 8 shows some of the correctly classified and misclassified characters by the proposed system.

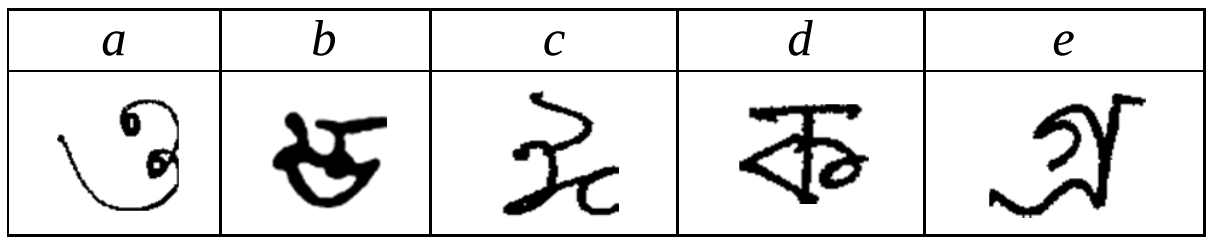

(a) Samples of correctly classified Bangla handwritten characters

| *Label* | *a* | *b* | *c* | *d* |
|---|---|---|---|---|
| *Misclassified character* | 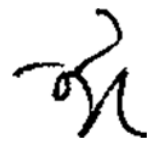 | 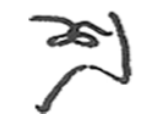 | 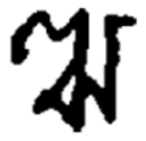 | 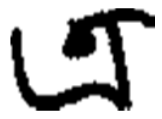 |
| *Misclassified character class* | 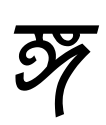 | 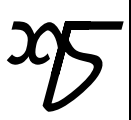 | 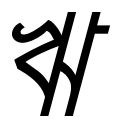 | 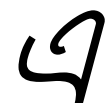 |

(b) Samples of misclassified Bangla handwritten characters with respective correct character class

**Figure 8**: Samples of correctly classified and misclassified characters by the proposed method

## V. CONCLUSION

An efficient system is proposed here for recognizing handwritten characters by identifying the regions of the character image which contain most of its discriminating features. An enhanced harmony search is used here for identifying the most informative regions. The proposed method is evaluated for three datasets. Evaluation is based on two factors, recognition accuracy and minimum number of local regions sufficient to identify the character correctly. From results of the experiments, the present work showed noticeable reduction in the number of most discriminating regions as well as significant increment of the recognition accuracy. To the best of our knowledge, this is the first work that uses the power of harmony search for sampling local regions to recognize handwritten characters. The results have shown great promise in this approach. Therefore it opens up a new frontier for more successful handwritten character recognition systems. Also it presents with future scope for researchers to improve its performance by using different feature-set or employing a more powerful variant of harmony search method present in the literature.

## ACKNOWLEDGEMENTS

Authors are thankful to *"Centre for Microprocessor Application for Training Education and Research"* and *Department of Computer Science & Engineering, Jadavpur University, Kolkata* for kindly providing infrastructural facilities that helped to complete this work.

## REFERENCES

[1] H. Fujisawa, "Forty years of research in character and document recognition—an industrial perspective," *Pattern Recognit.*, vol. 41, no. 8, pp. 2435–2446, Aug. 2008.

[2] U. Pal and B. B. Chaudhuri, "Indian script character recognition: a survey," *Pattern Recognit.*, vol. 37, no. 9, pp. 1887–1899, Sep. 2004.

[3] D. Impedovo and G. Pirlo, "Zoning methods for handwritten character recognition : A survey," *Pattern Recognit.*, vol. 47, no. 3, pp. 969–981, 2014.

[4] N. Das, S. Basu, R. Sarkar, M. Kundu, M. Nasipuri, and D. kumar Basu, "An Improved Feature Descriptor for Recognition of Handwritten Bangla Alphabet," Jan. 2015.

[5] N. Das, R. Sarkar, S. Basu, M. Kundu, M. Nasipuri, and D. K. Basu, "A genetic algorithm based region sampling for selection of local features in handwritten digit recognition application," *Appl. Soft Comput.*, vol. 12, no. 5, pp. 1592–1606, May 2012.

[6] A. Roy, N. Das, R. Sarkar, S. Basu, M. Kundu, and M. Nasipuri, "Region Selection in Handwritten Character Recognition using Artificial Bee Colony Optimization," pp. 183–186, 2012.

[7] M. Hanmandlu, A. V. Nath, A. C. Mishra, and V. K. Madasu, "Fuzzy Model Based Recognition of Handwritten Hindi Numerals using Bacterial Foraging," in *6th IEEE/ACIS International Conference on Computer and Information Science (ICIS 2007)*, 2007, pp. 309–314.

[8] G. V. Loganathan, "A New Heuristic Optimization Algorithm: Harmony Search," *Simulation*, vol. 76, no. 2, pp. 60–68, Feb. 2001.

[9] Z. W. Geem, "Particle-swarm harmony search for water network design," *Eng. Optim.*, vol. 41, no. 4, pp. 297–311, Apr. 2009.

[10] Z. W. Geem, K. S. Lee, and Y. Park, "Application of Harmony Search to Vehicle Routing," *Am. J. Appl. Sci.*, vol. 2, no. 12, pp. 1552–1557, Dec. 2005.

[11] Z. W. Geem, J. H. Kim, and G. V Loganathan, "Harmony search optimization: application to pipe network design," Jan. 2002.

[12] R. Sarkar, N. Das, S. Basu, M. Kundu, M. Nasipuri, and D. K. Basu, "CMATERdb1: a database of unconstrained handwritten Bangla and Bangla–English mixed script document image," *Int. J. Doc. Anal. Recognit.*, vol. 15, no. 1, pp. 71–83, Feb. 2011.

[13] "CMATERdb3.1.3.3 - cmaterdb - Handwritten Bangla Compound character image database - CMATERdb: The pattern recognition database repository - Google Project Hosting." [Online]. Available: https://code.google.com/p/cmaterdb/downloads/detail?name=CMATERdb3.1.3.3.7z&can=2&q=. [Accessed: 31-Jan-2015].

[14] S. Basu, N. Das, R. Sarkar, M. Kundu, M. Nasipuri, and D. K. Basu, "Recognition of numeric postal codes from multi-script postal address blocks," in *Pattern Recognition and Machine Intelligence*, Springer, 2009, pp. 381–386.

[15] K. S. Lee and Z. W. Geem, "A new meta-heuristic algorithm for continuous engineering optimization: harmony search theory and practice," *Comput. Methods Appl. Mech. Eng.*, vol. 194, no. 36–38, pp. 3902–3933, Sep. 2005.

[16] V. N. Vapnik, "An overview of statistical learning theory.," *IEEE Trans. Neural Netw.*, vol. 10, no. 5, pp. 988–99, Jan. 1999.

[17] C. Chang and C. Lin, "LIBSVM : A Library for Support Vector Machines," vol. 2, no. 3, 2011.

[18] N. Das, B. Das, R. Sarkar, S. Basu, and M. Kundu, "Handwritten Bangla Basic and Compound character recognition using MLP and SVM classifier," vol. 2, no. 2, pp. 109–115, 2010.